\documentclass[runningheads]{llncs}
\usepackage{graphicx}

\usepackage{tikz}
\usepackage{subfigure}
\usepackage{comment}
\usepackage{amsmath,amssymb} 
\usepackage{color}
\usepackage{url}
\usepackage{soul}
\usepackage{subfigure}
\usepackage{multirow}
\usepackage{booktabs}
\usepackage{xcolor,colortbl} 
\usepackage{setspace}
\usepackage{url}

\usepackage{algorithmic}
\usepackage[ruled,vlined]{algorithm2e}  

\definecolor{OliveGreen}{rgb}{0.0, 1.0, 0.0}
\usepackage[pagebackref=false,breaklinks=true,letterpaper=true,colorlinks,urlcolor = black,  citecolor = OliveGreen, bookmarks=false]{hyperref}

\usepackage[accsupp]{axessibility}  

\begin{document}
\pagestyle{headings}
\mainmatter
\def\ECCVSubNumber{1414}  

\title{Out-of-distribution Detection with \\ Boundary Aware Learning} 

\titlerunning{BAL}
%
\author{Sen Pei\inst{1} \and
Xin Zhang\inst{1} \and
Bin Fan\inst{4} \and
Gaofeng Meng \inst{1,2,3} \thanks{Corresponding author}}
\authorrunning{S. Pei et al.}
%
\institute{NLPR, Institute of Automation, Chinese Academy of Sciences, China \and
School of Artificial Intelligence, University of Chinese Academy of Sciences, China \and CAIR, HK Institute of Science and Innovation, China\and University of Science and Technology Beijing, China}

\maketitle

\begin{abstract}
There is an increasing need to determine whether inputs are out-of-distribution (\emph{OOD}) for safely deploying machine learning models in the open world scenario. Typical neural classifiers are based on the closed world assumption, where the training data and the test data are drawn \emph{i.i.d.} from the same distribution, and as a result, give over-confident predictions even faced with \emph{OOD} inputs. For tackling this problem, previous studies either use real outliers for training or generate synthetic \emph{OOD} data under strong assumptions, which are either costly or intractable to generalize. In this paper, we propose boundary aware learning (\textbf{BAL}), a novel framework that can learn the distribution of \emph{OOD} features adaptively. The key idea of BAL is to generate \emph{OOD} features from trivial to hard progressively with a generator, meanwhile, a discriminator is trained for distinguishing these synthetic \emph{OOD} features and in-distribution (\emph{ID}) features. Benefiting from the adversarial training scheme, the discriminator can well separate \emph{ID} and \emph{OOD} features, allowing more robust \emph{OOD} detection. The proposed BAL achieves \emph{state-of-the-art} performance on classification benchmarks, reducing up to 13.9\% FPR95 compared with previous methods. 
\end{abstract}

\section{Introduction}
Deep convolutional neural networks are one of the basic architectures in deep learning, and they have achieved great success in modern computer vision tasks. However, the over-confidence issue of \emph{OOD} data has always been with CNN which harms its generalization performance seriously. In previous researches, neural networks have been proved to generalize well when the test data is drawn \emph{i.i.d.} from the same distribution as the training data, i.e., the \emph{ID} data. However, when deep learning models are deployed in an open world scenario, the input samples can be \emph{OOD} data and therefore should be handled cautiously. 

\begin{figure}[t]
\centering
\subfigure[ResNet18~\cite{2}]{
\begin{minipage}[t]{0.33\linewidth}
\centering
\includegraphics[width=1.5in]{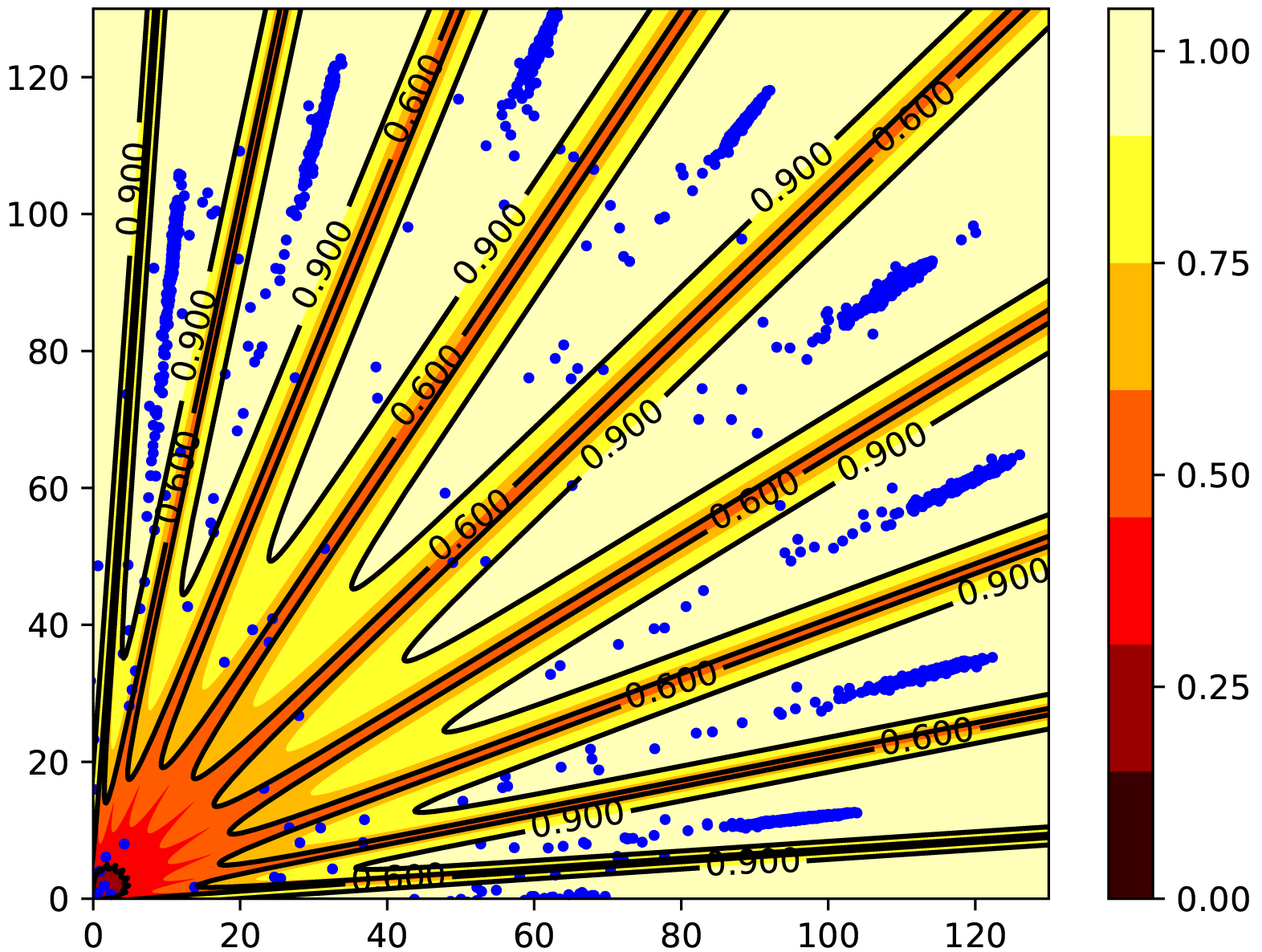}
\end{minipage}%
}%
\subfigure[MLP]{
\begin{minipage}[t]{0.33\linewidth}
\centering
\includegraphics[width=1.5in]{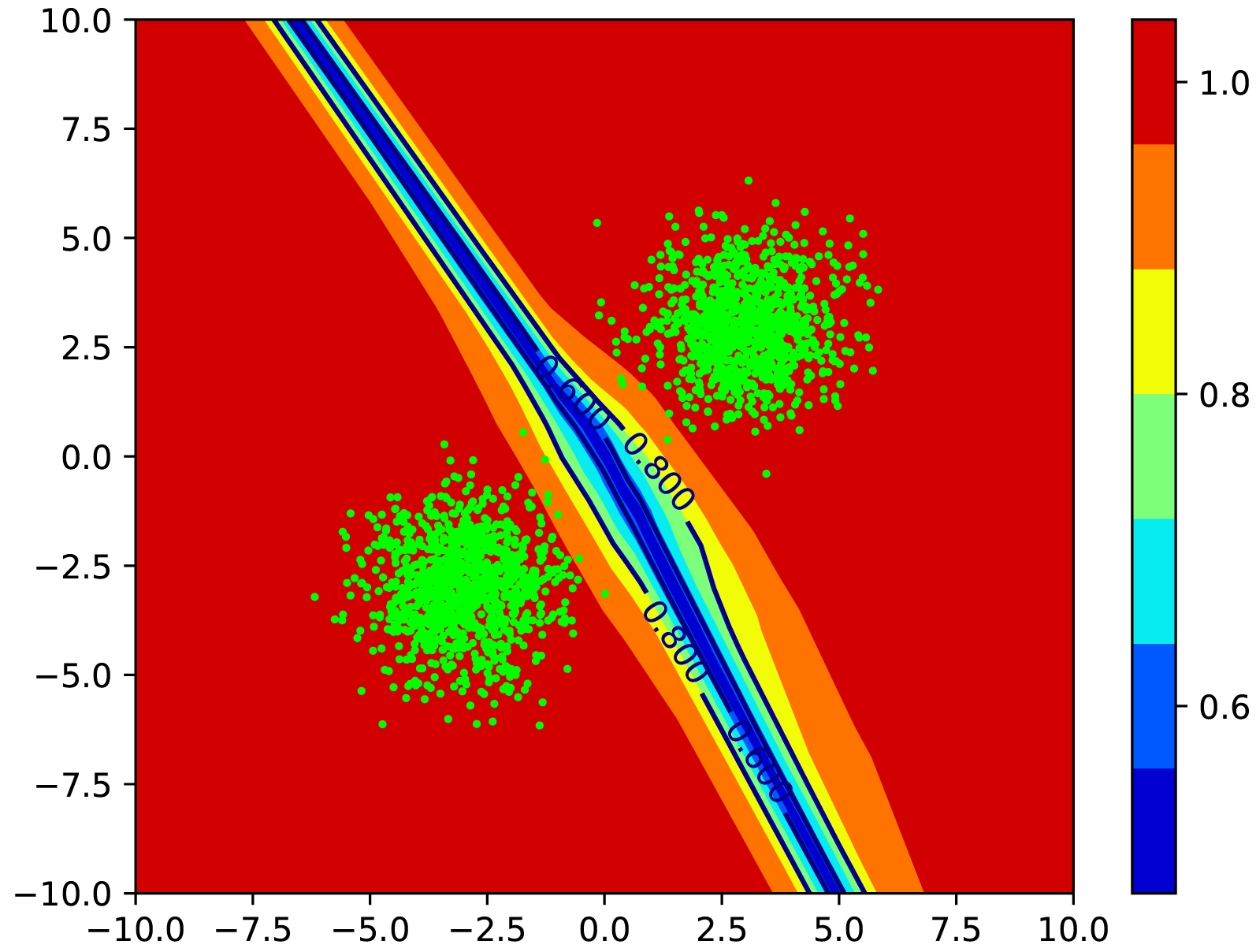}
\end{minipage}%
}%
\subfigure[BAL (ours)]{
\begin{minipage}[t]{0.33\linewidth}
\centering
\includegraphics[width=1.5in]{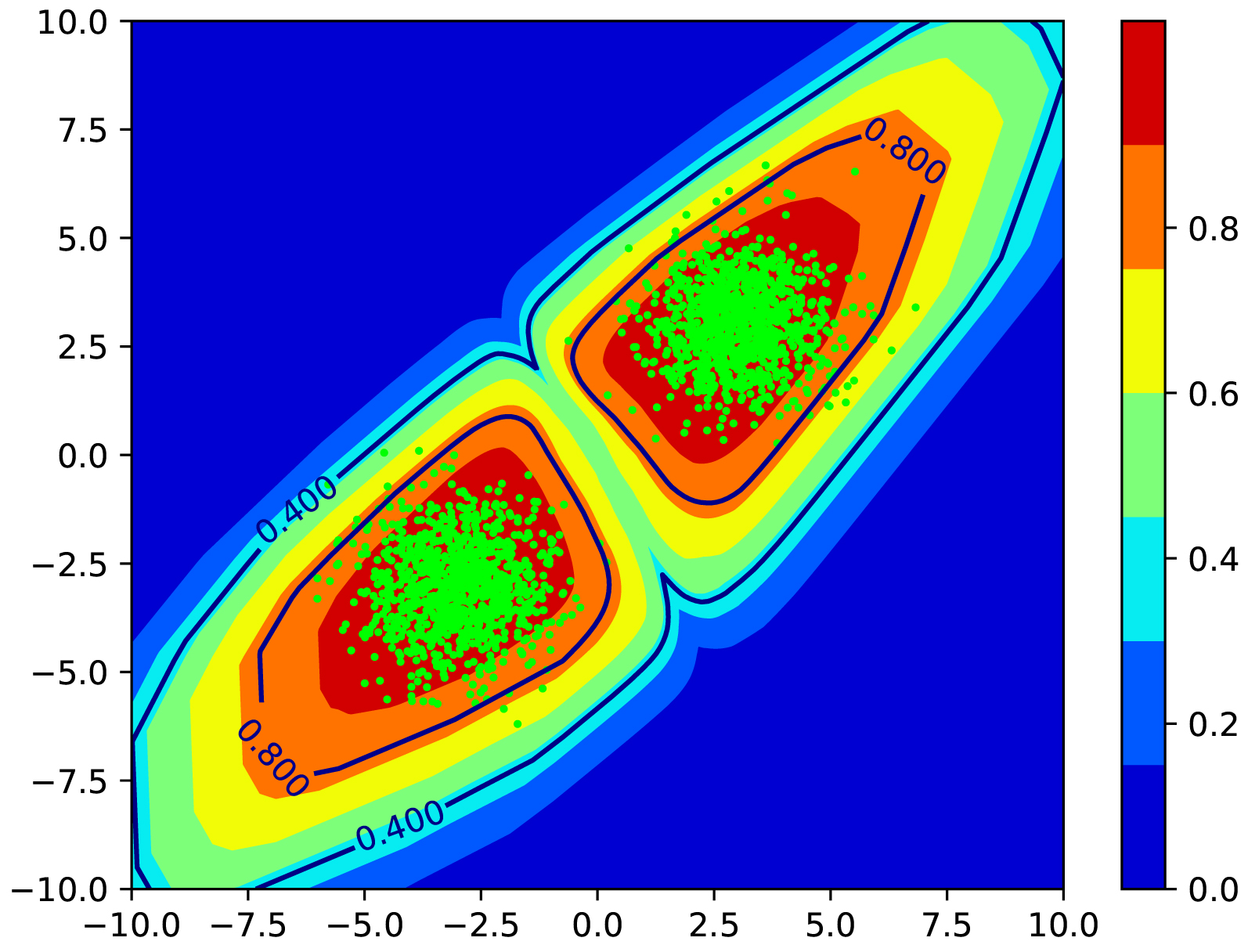}
\end{minipage}
}%
\centering
\vskip -0.15in
\caption{\textbf{Over-confidence issue in traditional classification nets.} (a): A ResNet18 trained on MNIST. The number of neurons of its penultimate layer is set to 2 for feature visualization. The \textcolor{blue}{\textbf{blue points}} are feature representations of \emph{ID} data. The background color represents confidence score given by the ResNet18. It is shown that the region far from the blue points gets high confidence score. (b): Classification on two gaussian distribution with a MLP. The \textcolor{green}{\textbf{green points}} are training data. It can be seen the classification net gives \emph{OOD} regions high confidence which is abnormal. (c): Boundary aware learning gives \emph{ID} regions much higher confidence than \emph{OOD} regions.}
\label{fig:illustration}
\vskip -0.3in
\end{figure}

Generally, there are two major challenges for improving the robustness of models: adversarial examples and \emph{OOD} examples. As pointed out in \cite{6}, adding very small perturbations to the input can fool a well-trained classification net, and these modified inputs are the so-called adversarial examples. Another problem is how to detect \emph{OOD} examples that are drawn far away from the training data. The trained neural networks often produce very high confidence to these \emph{OOD} samples which has raised concerns for AI Safety \cite{7} in many applications, which is the so-called over-confidence issue \cite{42}. As shown in Figure~\ref{fig:illustration} (a), a trained ResNet18 is used for extracting features from the MNIST dataset, and the blue points indicate feature representations of \emph{ID} data. It can be found that almost the whole feature space is assigned with high confidence score but the \emph{ID} data only concentrates in some narrow regions densely.

Previous studies have proposed different approaches for detecting \emph{OOD} samples to improve the robustness of classifiers. In \cite{8}, a max-softmax method is proposed for identifying \emph{OOD} samples. Further, in ODIN \cite{9}, temperature scaling and input pre-processing are introduced for improving the confidence scores of \emph{ID} samples. In \cite{10}, convolutional prototype learning is proposed for image classification which shows effectiveness in \emph{OOD} detection and class-incremental learning. In \cite{11}, it points out that the outputs of softmax can not represent the confidence of neural net actually, and thus, a new branch is separated for confidence estimation independently. All these previous works have brought many different perspectives and inspirations for solving the open world recognition tasks. However, these methods pay limited attention to the learning of \emph{OOD} features which is a key factor in \emph{OOD} detection. The neural networks can better detect \emph{OOD} samples if they are supervised by the \emph{trivial} and \emph{hard OOD} information, and that's why we argue \emph{OOD} feature learning is important for \emph{OOD} uncertainty estimation.

In this paper, we attribute the reason of poor \emph{OOD} detection performance to the fact that the traditional classification networks can not perceive the boundary of \emph{ID} data due to lack of \emph{OOD} supervision, as illustrated in Figure~\ref{fig:illustration} (a) and (b). Consequently, this paper focuses on how to generate synthetic \emph{OOD} information that supervises the learning of classifiers. The key idea of our proposed boundary aware learning (\textbf{BAL}) is to generate synthetic \emph{OOD} features from trivial to hard gradually via a generator. At the same time, a discriminator is trained to distinguish \emph{ID} and \emph{OOD} features. Powered by this adversarial training phase, the discriminator can well separate \emph{ID} and \emph{OOD} features. The key contributions of this work can be summarized as follows:
\begin{itemize}
\item A boundary aware learning framework is proposed for improving the rejection ability of neural networks while maintaining the classification performance. This framework can be combined with mainstream neural net architectures.

\item We use a GAN to learn the distribution of \emph{OOD} features adaptively step by step without introducing any assumptions about the distribution of \emph{ID} features. Alongside, we propose an efficient method called RSM (Representation Sampling Module) to sample synthetic \emph{hard OOD} features. 

\item We test the proposed BAL on several datasets with different neural net architectures, the results suggest that BAL significantly improves the performance of \emph{OOD} detection, achieving \emph{state-of-the-art} performance and allowing more robust classification in the open world scenario.

\end{itemize}

\section{Related Works}

\noindent
\textbf{\emph{OOD} detection with softmax-based scores.}\quad In \cite{8}, a baseline approach to detect \emph{OOD} inputs named max-softmax is proposed, and the metrics of evaluating \emph{OOD} detectors are defined properly. Following this, inspired by \cite{6}, ODIN \cite{9} and generalized ODIN \cite{50} are proposed for improving the detection ability of max-softmax using temperature scaling, input pre-processing and confidence decomposition. In \cite{13,14}, these studies argue that the feature maps from the penultimate layer of neural networks are not suitable for detecting outliers, and thus, they use the features from a well-chosen layer and adopt some metrics such as Euclidean distance, Mahalanobis distance and OSVM \cite{15}. In \cite{11}, a branch is separated out for confidence regression since the outputs of softmax can not well represent the confidence of neural networks. More recently, GradNorm \cite{51} finds that the magnitude of gradients is higher in \emph{ID} than that of \emph{OOD}, making it informative for \emph{OOD} detection. In \cite{52}, energy score derived from discriminative models is used for \emph{OOD} detection which also brings some improvement. 

\noindent
\textbf{\emph{OOD} detection with synthetic data.}\quad These kinds of methods usually use the \emph{ID} samples to generate fake \emph{OOD} samples, and then, train a $(C+1)$ classifier which can improve the rejection ability of neural nets. \cite{18} treats the \emph{OOD} samples as two types, one indicates these samples   that are close to but outside the \emph{ID} manifold, and the other is these samples which lie on the \emph{ID} boundary. This work uses Variational AutoEncoder \cite{46} to generate such data for training. In \cite{17}, the authors argue that samples lie on the boundary of \emph{ID} manifold can be treated as \emph{OOD} samples, and they use GAN \cite{20} to generate these data. The proposed joint training method of confident classifier and adversarial generator inspire our work. It can not be ignored that the methods mentioned above are only suitable for small toy datasets, and the joint training method harms the classification performance of neural nets. Further, in \cite{16}, the study points out that AutoEncoder can reconstruct the \emph{ID} samples with much less error than \emph{OOD} examples, allowing more effective detection with taking reconstruction error into consideration. Very recently, a newly proposed VOS \cite{49} introduces the \emph{OOD} detection into object detection tasks, and its main focus is still the \emph{OOD} feature generation. In these previous works, the features of each category from penultimate layer of CNN are assumed to follow a multivariate gaussian distribution. We argue and verify that this assumption is not reasonable. Our proposed BAL uses a GAN to learn the \emph{OOD} distribution adaptively without making assumptions, and the experimental results show that BAL outperforms gaussian assumption based methods significantly. 

\noindent
\textbf{Improving detection robustness with model ensembles.}\quad These kinds of methods are similar to bagging in machine learning. In \cite{23}, the authors initialize different parameters for neural networks randomly, and the bagging sampling method is used for generating training data. Similarly, in \cite{22}, the features from different layers of neural network are used for identifying \emph{OOD} samples. The defined higher order Gram Matrices in this work yield better \emph{OOD} detection performance. More recently, \cite{24} converts the labels of training data into different word embeddings using \emph{GloVe}~\cite{47} and \emph{FastText}~\cite{48} as the supervision to gain diversity and redundancy, the semantic structure improves the robustness of neural networks.

\noindent
\textbf{\emph{OOD} detection with auxiliary supervision.}\quad In \cite{38}, the authors argue that the likelihood score is heavily affected by the population level background statistics, and thus, they propose a likelihood ratio method to deal with background and semantic targets in image data. In \cite{39}, the study finds that self-supervision can benefit the robustness of recognition tasks in a variety of ways. In \cite{41}, a residual flow method is proposed for learning the distribution of feature space of a pre-trained deep neural network which can help to detect \emph{OOD} examples. The latest work in \cite{25} treats \emph{ood} samples as \emph{near-OOD} and \emph{far-OOD} samples, it argues that contrastive learning can capture much richer features which improve the performance in detecting \emph{near-OOD} samples. In \cite{40}, the author uses auxiliary datasets served as \emph{OOD} data for improving the anomaly detection ability of neural networks. Generally, these kinds of methods use some prior information to supervise the learning of \emph{OOD} detector. 
\vskip -0.2in
\section{Preliminaries}
\vskip -0.05in
\subsubsection{Problem Statement}
\begin{figure}[t]
\centering
\subfigure[Classification with ResNet18~\cite{2}]{
\begin{minipage}[t]{0.45\linewidth}
\centering
\includegraphics[width=2in]{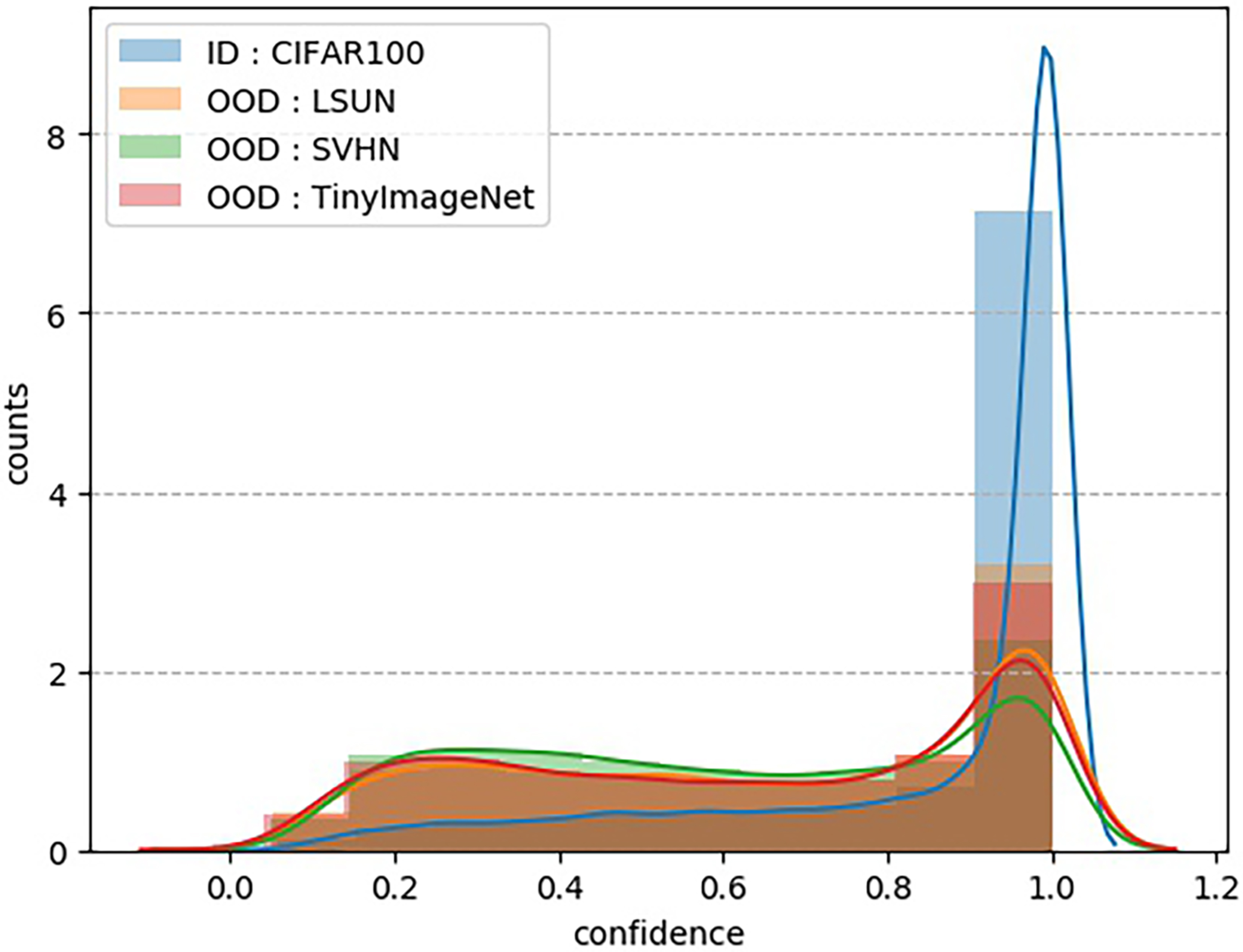}
\end{minipage}%
}%
\subfigure[Classification with BAL (ours)]{
\begin{minipage}[t]{0.45\linewidth}
\centering
\includegraphics[width=2in]{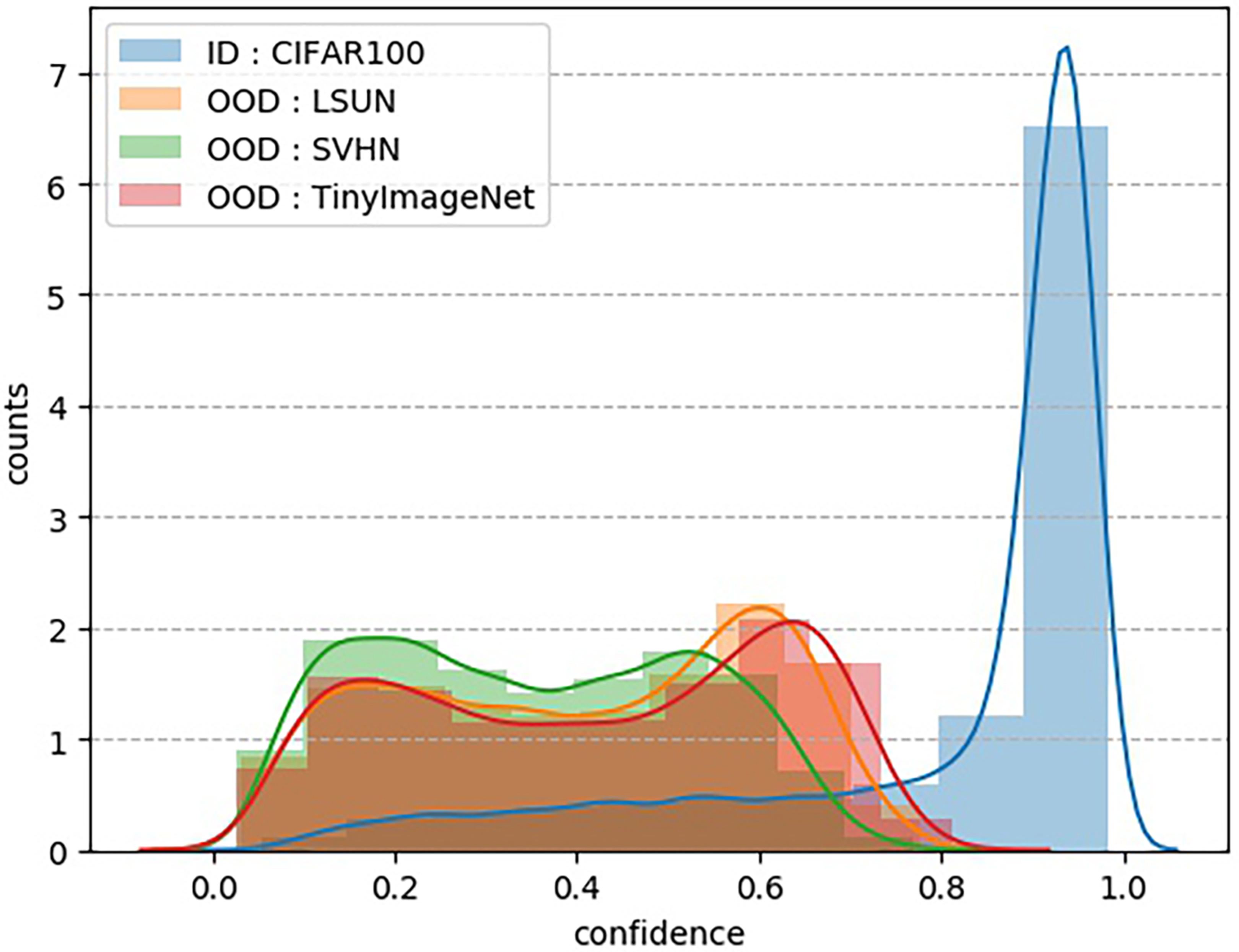}
\end{minipage}%
}%
\centering
\vskip -0.1in
\caption{\textbf{Confusion between \emph{ID} and \emph{OOD} data.} (a): In typical neural net, the \emph{ID} and \emph{OOD} data are confused, and both of them get very high confidence scores. (b): With proposed BAL, the \emph{OOD} data is assigned with much lower confidence, allowing more effective \emph{OOD} detection.}
\label{fig:idea}
\vskip -0.17in
\end{figure}

This work considers the problem of separating \emph{ID} and \emph{OOD} samples. Suppose $P_{in}$ and $P_{out}$ are distributions of \emph{ID} and \emph{OOD} data, $X=\{x_1,x_2,...,x_N\}$ are images randomly sampled from these two distributions. This task aims to give lower confidence of image $x_i$ sampled from $P_{out}$ while higher to that of $P_{in}$. Typically, \emph{OOD} detection can be formulated as a binary classification problem. With a chosen threshold $\gamma$ and confidence score $S(x)$, input is judged as \emph{OOD} data if $S(x) < \gamma$ otherwise \emph{ID}. Figure~\ref{fig:idea} (a) shows the traditional classifiers can not capture the \emph{OOD} uncertainty, and as a result, produce over-confident predictions on \emph{OOD} data. Figure~\ref{fig:idea} (b) shows an ideal case where \emph{ID} data gets higher score than \emph{OOD}. Methods that aim to boost the performance of \emph{OOD} detection should use no data labeled as \emph{OOD} explicitly.
\vskip -0.1in

\subsubsection{Methodology}
For a given image $x$, its corresponding feature representation $f$ can be got from the penultimate layer of a pre-trained classification net, and based on the total probability theorem, we have:
\begin{equation}
\begin{split}
P(w|f)=P(w|f\in \mathcal{M}_f)\cdot P(f\in \mathcal{M}_f|f)\\ + P(w|f\notin \mathcal{M}_f)\cdot P(f\notin \mathcal{M}_f|f)
\label{eq:total_p}
\end{split}
\end{equation}
where $w$ is the category label of \emph{ID} data, and $\mathcal{M}_f$ represents the manifold of \emph{ID} features. Typical neural networks have no access to \emph{OOD} data, therefore its softmax output is actually the conditional probability assuming the inputs are \emph{ID} data, i.e., $P(w|f\in \mathcal{M}_f)$. Empirically, since the \emph{OOD} data has quite different semantic meanings compared with \emph{ID} data, it is reasonable to approximate $P(w|f\notin \mathcal{M}_f)$ to 0. Then, we have:
\begin{equation}
P(w|f)\approx P(w|f\in \mathcal{M}_f)\cdot P(f\in \mathcal{M}_f|f)
\label{eq:proxy_p}
\end{equation}
It tells that the approximation of posterior can be formulated as the product of outputs from pre-trained classifiers and the probability $f$ belongs to $\mathcal{M}_f$. The proposed BAL aims to estimate $P(f\in \mathcal{M}_f|f)$ with features from the penultimate layer of pre-trained CNN.

\vskip -0.3in
\section{Boundary Aware Learning}
\begin{figure}[t]
\begin{center}
\centerline{\includegraphics[width=\columnwidth]{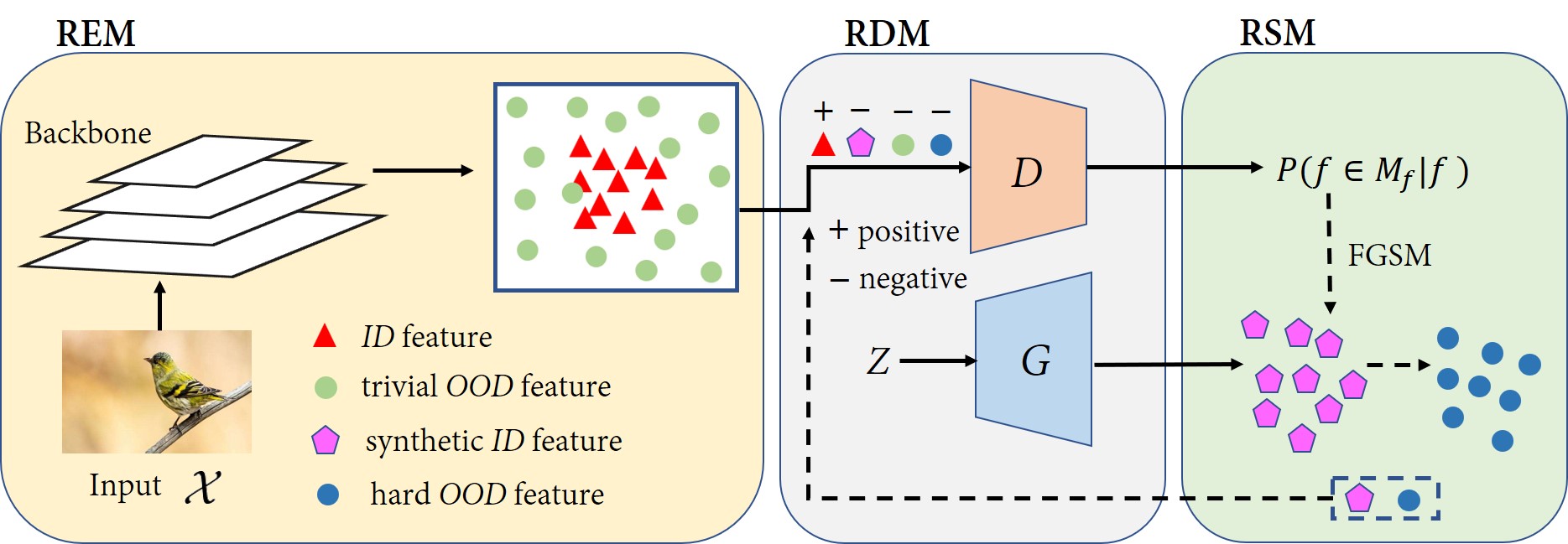}}
\vskip -0.1in
\caption{\textbf{The proposed BAL framework.} The \emph{ID} features are extracted from pre-trained classifier. The trivial \emph{OOD} features are uniformly sampled in feature space. The \emph{hard OOD} features are generated using fast gradient sign method (FGSM). In the process of training discriminator, all features except \emph{ID} feature are treated as \emph{OOD}. $\mathcal{M}_f$ is the manifold of \emph{ID} features. REM, RSM and RDM are representation extraction module, representation sampling module and representation discrimination module respectively.}
\label{framework}
\end{center}
\vskip -0.45in
\end{figure}

The proposed boundary aware learning framework contains three modules as illustrated in Figure~\ref{framework}. These modules handle the following problems: \textbf{(I)} Representation Extraction Module (REM) : how to generate trivial \emph{OOD} features to supervise the learning of conditional discriminator; \textbf{(II)} Representation Sampling Module (RSM) : how to generate synthetic \emph{hard OOD} features to enhance the discrimination ability of conditional discriminator step by step; \textbf{(III)} Representation Discrimination Module (RDM) : how to make the conditional discriminator aware the boundary of \emph{ID} features.
\vskip -0.1in

\subsection{Representation Extraction Module (REM)}
\label{sec:rem}
This module handles the problem of how to generate trivial synthetic \emph{OOD} features. As in prior works, we use the outputs of penultimate layer in CNN to represent the input images. In the following parts, $\mathcal{H}$ and $h$ are used to indicate the pre-trained classification net with and without the top classification layer, and $\theta$ is the pre-trained weights. Formally, the feature $f$ of an input image $x$ is described as:
\begin{equation}
f=h(x;\theta)
\label{eq:feat}
\end{equation}

During training, image $x$ and its corresponding label $c$ are sampled from dataset $\mathcal{X}$. We get an \emph{ID} feature-label pair $\left \langle f,c \right \rangle$ with Eq.(\ref{eq:feat}). For generating trivial synthetic \emph{OOD} features, we sample data in feature space uniformly. Given a batch features $\{f_1,f_2,f_3,...,f_k\}$, the length of each feature vector $f_i$ is $m$. We first calculate the minimal and maximal bound in $m$-dimensional space that contains all features within this batch. For $j\in \{1,2,3,...,m\}$, we have:
\vspace{-0.2em}
\begin{align}
R_{\min}^{(j)}&=\min_{1\leq i\leq k} f_{i}^{(j)} \label{eq:bound1}\\
R_{\max}^{(j)}&=\max_{1\leq i\leq k} f_{i}^{(j)} \label{eq:bound2}
\end{align}
therefore, the lower and upper bound of feature vectors are obtained as follows:
\vspace{-1em}
\begin{align}
a&=(R_{\min}^{(1)}, R_{\min}^{(2)},..., R_{\min}^{(m)})^T \label{eq:bound3}\\
b&=(R_{\max}^{(1)}, R_{\max}^{(2)},..., R_{\max}^{(m)})^T \label{eq:bound4}
\end{align}

\begin{figure}
\vskip -0.15in
\centering
\includegraphics[height=4cm]{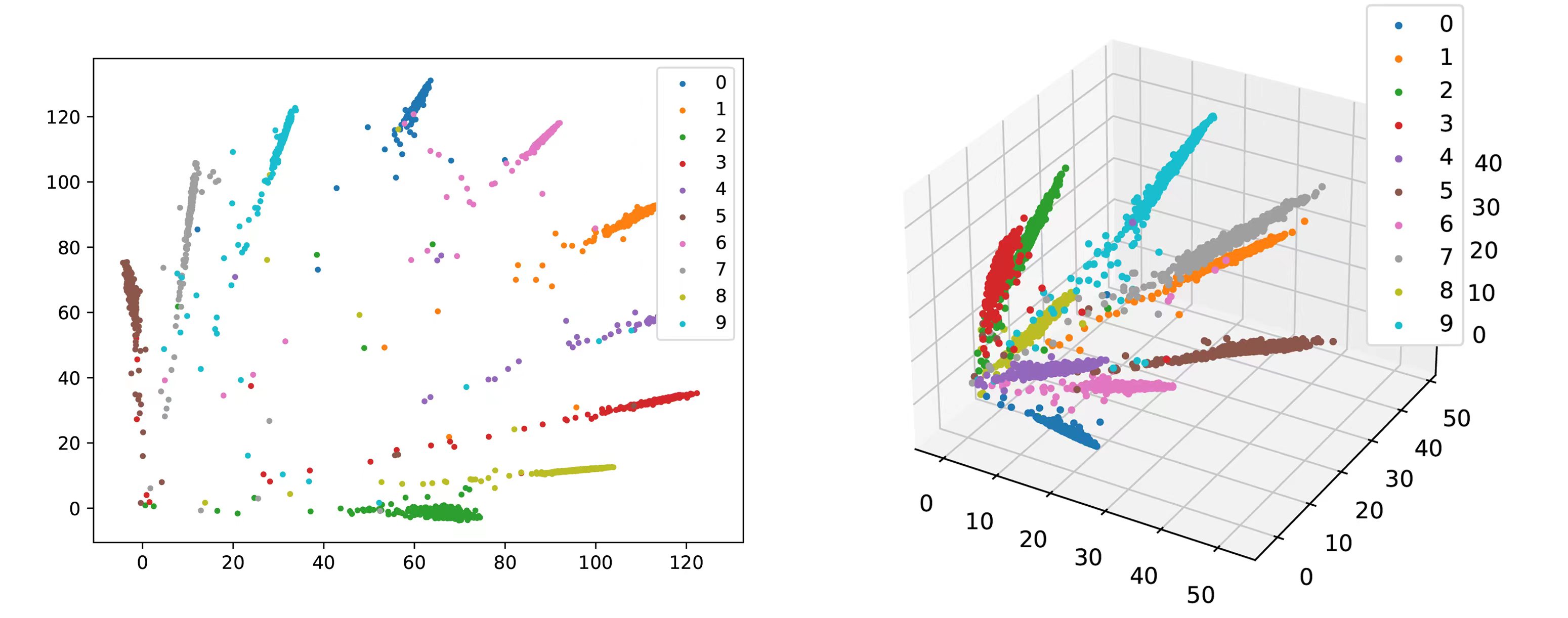}
\vskip -0.1in
\caption{\textbf{Feature distribution in penultimate layer of CNN.} Left: Classification on MNIST with ResNet18, the penultimate layer has 2 neuros for visualization. Right: Same as the left, the penultimate layer has 3 neurons. There is a large deviation between the distribution of \emph{ID} feature and a multivariate gaussian. Moreover, it is clear that \emph{ID} features densely distribute at some narrow regions in feature space.}
\label{dim_vis}
\vskip -0.45in
\end{figure}

We use $\mathbb{U}(a,b)$ to indicate a batch-wise uniform distribution in feature space. Randomly sampled feature $\hat{f}$ from $\mathbb{U}(a,b)$ is treated as negative sample with a randomly generated label $\hat{c}$. The negative pair is expressed as $\left \langle \hat{f},\hat{c} \right \rangle$. We give the reasons of uniform sampling : \textbf{(a)} It can not be guaranteed that features from the penultimate layer of CNN follow a multivariate gaussian distribution no matter in low dimensional space or higher feature space. For verifying this idea, we set the penultimate layer of CNN to have two and three neurons for feature visualization, the results shown in Figure~\ref{dim_vis} indicate the unreasonableness of this assumption. \textbf{(b)} \emph{ID} features densely distribute in some narrow regions which means the most samples from uniform sampling are \emph{OOD} data. 
Conflicts may happen when $\hat{f}$ is close to \emph{ID} and $\hat{c}$ does match with $\hat{f}$, the RDM deals with these conflicts.
\vskip -0.1in

\subsection{Representation Sampling Module (RSM)}
This module is used for generating \emph{hard OOD} features. For noise $z$ sampled from normal distribution $P_z$, its corresponding synthetic \emph{ID} feature $f$ can be got by $G(z,c)$ where $c$ is a conditional label. Since the generator $G$ is trained for generating \emph{ID} data, the feature $f$ is much closer to \emph{ID} instead of \emph{OOD}. With Fast Gradient Sign Method \cite{6}, we push the feature $f$ towards the boundary of \emph{ID} manifold which gets a much lower score from discriminator.
\begin{align}
\tilde{f}&=f-\epsilon\frac{\partial D(f;c)}{\partial f}\approx f-\epsilon \, \mathrm{sgn}(\frac{\partial D(f;c)}{\partial f}) \label{eq:fgsm1} \\
\tilde{z}&=z-\epsilon \frac{\partial D(f;c)}{\partial z}=z-\epsilon \frac{\partial D(G(z;c);c)}{\partial G(z;c)}\frac{\partial G(z;c)}{\partial z} \label{eq:fgsm2}
\end{align}
where $\tilde{f}$ represents the calibrated feature which scatters at the low density area of \emph{ID} feature distribution $P_f$. $\tilde{z}$ can be used for generating \emph{OOD} features by $G(\tilde{z};c)$. In particular, we set $\epsilon$ a random variable which follows a gaussian distribution for improving the diversity of sampling. $\left \langle \tilde{f},\tilde{c} \right \rangle$ is treated as \emph{hard OOD} feature pair because its quality is growing with the adversarial training process.

\vspace{-1em}
\subsection{Representation Discrimination Module (RDM)}
This module aims to make the discriminator aware the boundary of \emph{ID} features. The generator with FGSM is used for generating \emph{hard OOD} representations while the discriminator is used for separating \emph{ID} and \emph{OOD} features. The noise vector $z$ is sampled from a normal distribution $P_z$. The features of training images from REM follow a distribution $P_f$. For learning the boundary of \emph{ID} data via discriminator, we propose \textbf{shuffle loss} and \textbf{uniform loss}. The shuffle loss makes the discriminator aware the category of each \emph{ID} cluster in feature space, and the uniform loss makes the discriminator aware the boundary of each \emph{ID} feature cluster.
\vspace{-1em}
\subsubsection{Shuffle loss.} In each batch of the training data, we get feature-label pairs like $\left \langle f,c \right \rangle$. In a conditional GAN, these $\left \langle f,c \right \rangle$ pairs are treated as positive samples. With a shuffle function $T(\cdot)$, the positive pair $\left \langle f,c \right \rangle$ is transformed to a negative pair $\left \langle f,\tilde{c} \right \rangle$ where $\tilde{c}=T(c)\neq c$ is a mismatched label with feature $f$. The discriminator is expected to identify these mismatch pairs as \emph{OOD} data for awareness of category label, and the classification loss is the so called \textbf{shuffle loss} as below:
\begin{equation}
L_s = \mathbb{E}_{P_f} (\log D(f;T(c)) -\log D(f;c))
\label{eq:shuffle_loss}
\end{equation}

\vspace{-1.5em}
\subsubsection{Uniform loss.} We get positive pair $\left \langle f,c \right \rangle$ and negative pair $\left \langle \hat{f},\hat{c} \right \rangle$ from REM. It is mentioned before that conflicts may happen when $\hat{f}$ is close to some \emph{ID} feature clusters and the randomly generated label $\hat{c}$ dose match with them. For tackling this issue, we strengthen the memory of discriminator about positive pair $\left \langle f,c \right \rangle$ while weaken that about negative pair $\left \langle \hat{f},\hat{c} \right \rangle$. We force the discriminator to maximize $D(f;c)$ for remembering positive pairs, meanwhile, a hyperparameter $\lambda_c$ is used to mitigate the negative effects of conflicts. The \textbf{uniform loss} is defined as follows:
\begin{equation}
L_u = \lambda_c \cdot \mathbb{E}_{P_{\mathbb{U}}} \log D(\hat{f};\hat{c}) - \mathbb{E}_{P_f}\log D(f;c)
\label{eq:uniform_loss}
\end{equation}

Alongside, the \emph{hard OOD} features from RSM introduce no conflicts, and they are treated as negative \emph{OOD} pairs for calculating uniform loss when training discriminator. Formally, the loss function $L_d$ for conditional discriminator can be formulated as below:
\begin{align}
L_t &=-\mathbb{E}_{P_f}\log D(f;c)-\mathbb{E}_{P_z} \log(1-D(G(z);c)) \label{eq:d_loss1}\\
L_d &= L_t + L_s + L_u \label{eq:d_loss2}
\end{align}
where $L_t$ is the loss of discriminator in a vanilla conditional GAN. A well trained discriminator is a binary classifier for separating \emph{ID} and \emph{OOD} features. In the process of training generator, we add a regularization term to accelerate the convergence. The loss function of generator is written as:
\begin{equation}
L_g = \mathbb{E}_{P_z} \log(1-D(G(z;c);c))+ \lambda(\min_{f_c\in \mathcal{M}_c}||f_c-G(z;c)||_1)
\label{eq:generator_loss}
\end{equation}
where $||\cdot||_1$ indicates the L1 norm, $\mathcal{M}_c$ is the set of \emph{ID} features with label $c$, and $\lambda$ is a balance hyperparameter. The regularization term reduces the difference between synthetic features and the real. We set $\lambda$ to 0.01 in our experiments. In the process of training generator, the label $c$ is generated randomly.

Generally, the BAL framework only trains the conditional GAN while keeps the pre-trained classification net unchanged. The confidence score outputted by a trained discriminator is treated as $P(f\in \mathcal{M}_f|f)$. Based on Eq.(\ref{eq:proxy_p}), the approximation of posteriori is formulated as the product of outputs from pre-trained classification net and discriminator. The whole training and inference pipeline is shown in \textbf{Algorithm~\ref{bal}}.
\vskip -0.2in
\begin{algorithm}[htpb]
\label{bal}
    \caption{\emph{OOD} Detection with Boundary Aware Learning}
    \LinesNumbered
    \KwIn{pre-trained network $\mathcal{H}$(backbone $h$) on \emph{ID} data with parameter $\theta$, initial generator $G$, initial discriminator $D$, \emph{ID} dataset $\mathcal{X}$ }
    \KwOut{\emph{OOD} discriminator $D$, synthetic \emph{ID} generator $G$}
    \While {\texttt{Training}}{
		\For {\texttt{iter} = 1 to M}{
			\texttt{\# Discriminator training}\;
			{Sample a batch data \textbf{$x$} from \textbf{$\mathcal{X}$}\;
			Get the corresponding feature vectors : $f=h(x;\theta)$\;
			Calculate the lower and upper bound of $f$ with Eqs.(\ref{eq:bound1},\ref{eq:bound2},\ref{eq:bound3},\ref{eq:bound4})\;
			Transform the positive pairs $\left \langle f,c \right \rangle$ into negative pairs $\left \langle f,T(c) \right \rangle$\;
			Sample \emph{trivial OOD} and \emph{hard OOD} feature pairs $\left \langle \hat{f},\hat{c} \right \rangle$ via uniform sampling and RSM\;
			Calculate the shuffle loss $L_s$, the uniform loss $L_u$, and the vanilla loss $L_t$ with Eqs.(\ref{eq:shuffle_loss},\ref{eq:uniform_loss},\ref{eq:d_loss1})\;
			Update the parameters of $D$ with gradient descent method.}

			\texttt{\# Generator training}\;
			{Sample noise $z$ from normal distribution\;
			Get the features conditioned by random labels : $G(z;c)$\;
			Calculate the loss function of generator with Eq.(\ref{eq:generator_loss})\;
			Update the parameters of $G$ with gradient descent method.}
}}

    \While {\texttt{Inference}}{
			Get feature vector : $\hat{f} = h(\hat{x};\theta)$\;
			Get predict label and corresponding confidence: $p_1,\hat{c} = \mathcal{H}(\hat{x};\theta)$\;
			Get \emph{ID} confidence score : $p_2 = D(\hat{f},\hat{c})$\;
			Perform \emph{OOD} detection with $p_1\cdot p_2$ under a chosen threshold.
}
\end{algorithm}

\vspace{-3em}
\section{Experiments}
\vskip -0.05in
In this section, we validate the proposed BAL framework on several image classification datasets and neural net architectures. Experimental setup is described from Section~\ref{sec:data} to Section~\ref{sec:setup}, ablation study is described in Section~\ref{sec:ablation}. We report the main results and metrics in Section~\ref{sec:main_results}. Visualization of synthetic \emph{OOD} data is given in Section~\ref{sec:vis}.
\vspace{-1em}

\subsection{Dataset}
\label{sec:data}
\noindent
\textbf{MNIST} \cite{27} : A database of handwritten digits in total 10 categories, has a training set of 60k examples, and a test set of 10k examples.

\noindent
\textbf{Fashion-MNIST} \cite{30} : A dataset contains grayscale images of fashion products from 10 categories, has a training set of 60k images, and a test set of 10k images.  

\noindent
\textbf{Omniglot} \cite{31} : A dataset that contains 1623 different handwritten characters from 50 different alphabets. In this work, we treat Omniglot as \emph{OOD} data.

\noindent
\textbf{CIFAR-10 and CIFAR-100} \cite{26} : The former one contains 60k colour images in 10 classes, with 6k images per class. The latter one also contains 60k images but in 100 classes, with 600 images per class. 

\noindent
\textbf{TinyImageNet} \cite{32} : A dataset contains 120k colour images in 200 classes, with 600 images per class.

\noindent
\textbf{SVHN} \cite{33} and \textbf{LSUN} \cite{34} : The former one contains colour images of street view house number. The latter one is a large-scale scene understanding dataset.
\vspace{-2em}
\subsection{Evaluation metrics}
We report the following metrics to measure the performance of \emph{OOD} detection. The quantity of \emph{ID} and \emph{OOD} examples are strictly kept same in evaluation.
\noindent
\textbf{FPR at 95\% TPR (FPR95)} is the probability of an \emph{OOD} example being misclassified as \emph{ID} examples when the True Positive Rate is 95\%. True positive Rate and False Positive Rate are the same as defined in \emph{ROC} curve. 

\noindent
\textbf{Detection Error} measures the misclassification probability when True Positive Rate is 95\%. It is defined as $0.5(1-\mathrm{TPR})+0.5\mathrm{FPR}$.

\noindent
\textbf{AUROC} represents the area under \emph{ROC} curve. Greater AUROC indicates that the neural network is more confident to assign higher score to \emph{ID} data than \emph{OOD} data. An ideal classifier has an AUROC score of 100\%.

\noindent
\textbf{AUPR} represents the area under Precision-Recall curve. AUPR$_{in}$ indicates the ability of detecting \emph{ID} data while AUPR$_{out}$ indicates that of \emph{OOD} data.

\vspace{-1em}
\subsection{Experimental setup}
\label{sec:setup} 

\noindent
\textbf{Softmax baseline.} ResNet~\cite{2} and DenseNet~\cite{3} are used as backbones, and they are trained with an Adam optimizer using cross-entropy loss in total of 300 epochs. Images from MNIST, Fashion-MNIST and Omniglot are resized to 28 $\times$ 28 with only one channel. Other datasets are resized to 32 $\times$ 32 with RGB channels. For MNIST, Fashion-MNIST and Omniglot, ResNet18 is used as the feature extractor. For any other datasets, ResNet34 and DenseNet-BC with 100 layers are used for feature extraction. 

\noindent
\textbf{GCPL.} We use distance-based cross-entropy loss and prototype loss as mentioned in \cite{10} for generalized convolutional prototype learning. The hyperparameter $\lambda$ that controls the weight of prototype loss is set to 0.01.

\noindent
\textbf{ODIN} and \textbf{Generalized ODIN.} For ODIN, parameters ($T$, $\epsilon$) are (10, 2e-3), (100, 5e-4) and (10, 5e-4) for CIFAR-10, CIFAR-100 and SVHN respectively.

\noindent
\textbf{AEC.} This method uses reconstruction error to detect outliers. The architectures and experimental details can be found in \cite{16}, we use the same pipeline to reproduce it.

\vspace{-0.7em}
\subsection{Ablation study}
\vspace{-0.1em}
\label{sec:ablation}
\subsubsection{Ablation on proposed loss functions.}
We compare different loss functions proposed in BAL. Specifically, we use DenseNet-BC as the feature extractor. CIFAR-10 is set as \emph{ID} data while TinyImageNet is set as \emph{OOD} data. We consider four combinations of proposed loss functions: $L_t$, $L_t+L_s$, $L_t+L_u$ and $L_t+L_s+L_u$. The details of pre-mentioned loss functions can be found in Eqs.(\ref{eq:shuffle_loss}-\ref{eq:d_loss1}). For uniform loss $L_u$, we set the hyperparameter $\lambda_c$ to 0.7. The results are summarized in Table~\ref{tab:table3}, where BAL with shuffle loss and uniform loss outperforms the alternative combinations. Compared to max-softmax, BAL reduces FPR95 up to 36.1\%.

\setlength{\tabcolsep}{3pt}
\begin{table}[htpb]
\begin{center}
\vskip -0.15in
\caption{Ablation on different combinations of loss functions. All networks are trained with the training set of CIFAR-10, and \textbf{no} \emph{OOD} data is used. $\lambda_c$ in uniform loss $L_u$ is set to 0.7. It can be seen that the proposed shuffle loss and uniform loss enhance the ability for detecting outliers.}
\label{tab:table3}
\begin{tabular}{@{}lcccc@{}}
\toprule
             & $\uparrow$ AUPR$_{in}$ & $\uparrow$ AUPR$_{out}$ & $\uparrow$ AUROC & $\downarrow$ FPR 95\\ \midrule
Softmax baseline     & 95.3    & 92.2     & 94.1 & 41.1 \\
BAL ($L_t$)         & 97.0    & 96.0     & 96.6 & 17.9\\
BAL ($L_t+L_s$) & 97.1    & 96.2     & 96.6  & 9.3\\
BAL ($L_t+L_u$) & 97.2    & 96.3     & 96.7 & 8.1\\ \rowcolor{gray!15}
\textbf{BAL ($L_t+L_s+L_u$)}         & \textbf{98.2}    & \textbf{98.0}     & \textbf{97.0}  & \textbf{5.0}\\ \bottomrule
\end{tabular}
\end{center}
\vskip -0.55in
\end{table}

\subsubsection{Ablation on $\lambda_c$ in uniform loss.}
We consider different value settings to test the sensitivity of $\lambda_c$ in Eq.(\ref{eq:uniform_loss}). CIFAR-10 and TinyImageNet are set as \emph{ID} and \emph{OOD} respectively. DenseNet-BC is used as the backbone. The ablation results shown in Table~\ref{tab:table4} demonstrate that with the increasing of $\lambda_c$, AUPR$_{out}$ of neural networks increases synchronously which means the classifier can aware more \emph{OOD} data. In particular, using $\lambda_c$ as 0.7 yields both better \emph{ID} and \emph{OOD} detection performance.

\setlength{\tabcolsep}{3.5pt}
\begin{table}[htpb]
\begin{center}
\vskip -0.1in
\caption{Ablation on parameter $\lambda_c$. All networks are trained with the training set of CIFAR-10, and \textbf{no} \emph{OOD} data is used. In the following experiments, if not specified, $\lambda_c$ is set to 0.7 throughout.}
\label{tab:table4}
\begin{tabular}{cccc>{\columncolor{gray!15}}cc}
\toprule
$\lambda_c$        & 0.1  & 0.3  & 0.5  & 0.7  & 0.9  \\ \hline
AUROC    & 94.8 & 95.2 & 96.7 & \textbf{97.0} & 96.2 \\
AUPR$_{in}$  & 95.3 & 95.3 & 97.1 & \textbf{98.2} & 96.3 \\
AUPR$_{out}$ & 92.1 & 93.4 & 96.9 & \textbf{98.0} & 97.1 \\ \bottomrule
\end{tabular}
\end{center}
\vskip -0.3in
\end{table}

\vspace{-1em}
\subsubsection{Ablation on \emph{OOD} synthesis sampling methods.}
We consider different trivial \emph{OOD} feature sampling methods. As described in Section~\ref{sec:rem}, the distribution of features in convolutional layer are usually assumed to follow a multivariate gaussian distribution. Therefore, the low density area of each category is treated as \emph{OOD} region. We argue this assumption is not reasonable enough because : \textbf{(I)} From Figure~\ref{fig:illustration} (a) and Figure~\ref{dim_vis}, we can see that in low dimensional feature space, the conditional distribution of each category has a great deviation with multivariate gaussian distribution; \textbf{(II)} In high dimensional space, the distribution of \emph{ID} features is extremely sparse, therefore it is hard to estimate the probability density of assumed gaussian distribution accurately; \textbf{(III)} It is costly to calculate the mean vector $\mu$ and covariance matrix $\Sigma$ of multivariate gaussian distribution in feature space with high dimensionality; \textbf{(IV)} Inefficient sampling. It is in low efficiency since the probability density need to be calculated for each synthetic sample. 

Without introducing any strong assumptions about the \emph{ID} features, we verify that the naive uniform sampling together with a GAN framework can model the \emph{OOD} feature distribution effectively. We still use CIFAR-10 and TinyImageNet as \emph{ID} and \emph{OOD} data. We compare uniform sampling and gaussian sampling in feature space. The dimensionality of features is controlled by setting different number of neurons in the penultimate fully connected layer. The ablation results are shown in Table~\ref{tab:table5}. It is clear that BAL with uniform sampling outperforms gaussian sampling in both low and high dimensional feature space.

\setlength{\tabcolsep}{4.5pt}
\begin{table}[htpb]
\begin{center}
\vskip -0.13in
\caption{Ablation on BAL with different sampling methods. The values in the table are AUROC. Both uniform sampling and gaussian sampling are performed within BAL framework.}
\label{tab:table5}
\begin{tabular}{cccccc}
\toprule
feature dim & 2    & 64   & 256  & 512  & 1024 \\ \hline 
BAL (Gaussian) & 94.3 & 96.4 & 96.9 & 98.1 & 98.5 \\  \rowcolor{gray!15}
BAL (Uniform)  & 96.5 & 97.0 & 97.3 & 98.1 & \textbf{98.8} \\ \bottomrule
\end{tabular}
\end{center}
\vskip -0.55in
\end{table}

\subsection{Detection results}
\vspace{-0.2em}
\label{sec:main_results}
We detail the main experimental results on several datasets with ResNet18, ResNet34 and DenseNet-BC. For CIFAR-10, CIFAR-100 and SVHN, we use the pre-trained ResNet-34 and DenseNet-BC, and for MNIST, Fashion-MNIST and Omniglot, we train the ResNet18 from scratch.

\vspace {-1em}
\subsubsection{Results on MNIST, Fashion-MNIST and Omniglot.}
We observe the effects of BAL in two groups. In the first group, MNIST is \emph{ID} data, and the mixture of Fashion-MNIST and Omniglot is \emph{OOD} data. In the second group, Fashion-MNIST is \emph{ID} data while MNIST and Omniglot are \emph{OOD} data. For simplicity, Cls Acc and Det Err are used to represent Classification Accuracy and Detection Error. For ODIN, temperature ($T$) and magnitude ($\epsilon$) are 10 and 5e-4 respectively. The results summarized in Table~\ref{tab:table1} tell that BAL is effective on image classification benchmark, particularly, BAL reduces FPR95 up to 24.1\% compared with ODIN in the second group.

\setlength{\tabcolsep}{5pt}
\begin{table}[t]\footnotesize
\begin{center}
\caption{Detecting \emph{OOD} samples on MNIST, Fashion-MNIST and Omniglot with ResNet18. We use the mixture of two datasets as \emph{OOD} samples.}
\vskip -0.07in
\label{tab:table1}
\begin{tabular}{lcccccccc}
\toprule
\emph{ID}       & \multicolumn{4}{c|}{MNIST}                         & \multicolumn{4}{c}{F-MNIST}           \\
\emph{OOD}      & \multicolumn{4}{c|}{F-MNIST \& Omniglot}           & \multicolumn{4}{c}{MNIST \& Omniglot} \\ \hline
Methods  & \multicolumn{8}{c}{Softmax baseline\cite{8} / ODIN\cite{9} / GCPL\cite{10} / \textbf{BAL(ours)}}                                   \\ \hline
$\uparrow$ Cls Acc  & \textbf{99.43} & \textbf{99.43} & 99.23 & \multicolumn{1}{c|}{\textbf{99.43}} & \textbf{91.51}   & \textbf{91.51}   & 90.93   & \textbf{91.51}   \\
$\downarrow$ Det Err  & 4.14  & 5.01  & 4.77  & \multicolumn{1}{c|}{\textbf{3.06}}  & 32.42   & 19.14   & 30.73   & \textbf{7.10}    \\
$\downarrow$ FPR 95 & 3.29  & 5.03  & 4.54  & \multicolumn{1}{c|}{\textbf{1.11}}  & 59.84   & 33.27   & 56.45   & \textbf{9.20}    \\
$\uparrow$ AUROC    & 97.66 & 97.94 & 97.96 & \multicolumn{1}{c|}{\textbf{99.32}} & 89.44   & 93.45   & 81.79   & \textbf{97.82}   \\
$\uparrow$ AUPR$_{in}$   & 97.22 & 97.42 & 98.14 & \multicolumn{1}{c|}{\textbf{99.46}} & 90.80   & 94.28   & 72.40   & \textbf{98.31}   \\
$\uparrow$ AUPR$_{out}$  & 97.24 & 97.64 & 97.35 & \multicolumn{1}{c|}{\textbf{99.09}} & 86.20   & 91.36   & 82.38   & \textbf{96.95}   \\ \bottomrule
\end{tabular}
\end{center}
\vskip -0.16in
\end{table}

\setlength{\tabcolsep}{2.3pt}
\begin{table}[t]\scriptsize
\begin{spacing}{1.19}
\begin{center}
\caption{Main \emph{OOD} detection results. We use C-10, C-100, TIN, D-BC and R-34 to represent CIFAR-10, CIFAR-100, TinyImageNet, DenseNet-BC and ResNet-34.}
\vskip -0.13in
\label{tab:table2}
\begin{tabular}{ccccccccccccccccc}
\toprule
ID                                                                    & OOD  & \multicolumn{5}{c}{\multirow{2}{*}{\begin{tabular}[c]{@{}c@{}}$\downarrow$ FPR\\ at 95\% TPR\end{tabular}}} & \multicolumn{5}{c}{\multirow{2}{*}{\begin{tabular}[c]{@{}c@{}}$\uparrow$ AUPR\\ in\end{tabular}}} & \multicolumn{5}{c}{\multirow{2}{*}{\begin{tabular}[c]{@{}c@{}}$\uparrow$ AUPR\\ out\end{tabular}}} \\
                                                                      &      & \multicolumn{5}{c}{}                                                                             & \multicolumn{5}{c}{}                                                                   & \multicolumn{5}{c}{}                                                                    \\ \hline
                                                                      &      & \multicolumn{15}{c}{Softmax baseline\cite{8} / AEC\cite{16} / ODIN\cite{9} / Generalized ODIN\cite{50} / \textbf{BAL(ours)}}                                                                                                                                                                                                   \\ \cline{3-17} 
\multirow{3}{*}{\begin{tabular}[c]{@{}c@{}}C-10\\ D-BC\end{tabular}}  & SVHN & 59.8          & 57.2          & 63.6          & 44.2         & \multicolumn{1}{c|}{\textbf{32.6}}         & 91.9        & 92.3        & 89.1        & 94.6       & \multicolumn{1}{c|}{\textbf{99.7}}       & 87.0            & 92.5            & 83.9            & 88.7            & \textbf{99.7}            \\
                                                                      & LSUN & 33.4          & 27.6          & 5.6           & 5.2          & \multicolumn{1}{c|}{\textbf{4.7}}          & 96.4        & 97.3        & 98.9        & 99.0       & \multicolumn{1}{c|}{\textbf{99.5}}       & 94.0            & 96.3            & 98.7            & \textbf{98.9}            & \textbf{98.9}            \\
                                                                      & TIN  & 41.1          & 35.1          & 10.5          & 9.3          & \multicolumn{1}{c|}{\textbf{5.0}}          & 95.3        & 96.2        & 98.1        & 97.9       & \multicolumn{1}{c|}{\textbf{98.2}}       & 92.2            & 94.0            & 97.8            & 97.4            & \textbf{98.0}            \\ \hline
\multirow{3}{*}{\begin{tabular}[c]{@{}c@{}}C-10\\ R-34\end{tabular}}  & SVHN & 67.5          & 57.2          & 64.4          & 12.7         & \multicolumn{1}{c|}{\textbf{11.3}}         & 92.2        & 93.4        & 85.8        & 94.5       & \multicolumn{1}{c|}{\textbf{95.5}}       & 84.9            & 84.5            & 81.8            & 93.4            & \textbf{97.4}            \\
                                                                      & LSUN & 54.6          & 34.6          & 26.2          & 21.3         & \multicolumn{1}{c|}{\textbf{15.8}}         & 92.3        & 91.8        & 93.7        & \textbf{94.0}       & \multicolumn{1}{c|}{93.9}       & 88.5            & 92.1            & 93.8            & 93.9            & \textbf{94.1}            \\
                                                                      & TIN  & 55.3          & 28.7          & 28.0          & 27.4         & \multicolumn{1}{c|}{\textbf{21.6}}         & 92.4        & 93.1        & 94.0        & \textbf{94.3}       & \multicolumn{1}{c|}{93.9}       & 88.3            & 90.1            & 92.9            & 92.7            & \textbf{93.8}            \\ \hline
\multirow{3}{*}{\begin{tabular}[c]{@{}c@{}}C-100\\ D-BC\end{tabular}} & SVHN & 73.3          & 63.2          & 60.9          & 31.9         & \multicolumn{1}{c|}{\textbf{21.5}}         & 85.9        & 89.3        & 90.2        & 90.7       & \multicolumn{1}{c|}{\textbf{91.5}}       & 78.5            & 86.7            & 85.2            & 89.5            & \textbf{92.8}            \\
                                                                      & LSUN & 83.3          & 66.0          & 58.4          & 23.9         & \multicolumn{1}{c|}{\textbf{11.3}}         & 72.4        & 87.4        & 85.0        & 88.1       & \multicolumn{1}{c|}{\textbf{89.3}}       & 65.4            & 84.9            & 82.0            & 87.6            & \textbf{88.7}            \\
                                                                      & TIN  & 82.4          & 59.7          & 56.9          & 22.7         & \multicolumn{1}{c|}{\textbf{12.0}}         & 73.0        & 83.7        & 84.7        & 86.5       & \multicolumn{1}{c|}{\textbf{91.5}}       & 67.4            & 82.9            & 83.0            & 84.3            & \textbf{90.6}            \\ \hline
\multirow{3}{*}{\begin{tabular}[c]{@{}c@{}}C-100\\ R-34\end{tabular}} & SVHN & 79.7          & 76.5          & 76.5          & 31.2         & \multicolumn{1}{c|}{\textbf{17.3}}         & 81.5        & 82.5        & 73.8        & 85.3       & \multicolumn{1}{c|}{\textbf{87.1}}       & 74.5            & 79.6            & 74.2            & 85.1            & \textbf{89.3}            \\
                                                                      & LSUN & 81.2          & 52.1          & 54.6          & 27.1         & \multicolumn{1}{c|}{\textbf{18.7}}         & 76.0        & 80.0        & 82.4        & 89.0       & \multicolumn{1}{c|}{\textbf{91.5}}       & 70.1            & 78.4            & 84.1            & \textbf{89.0}            & 88.7            \\
                                                                      & TIN  & 79.6          & 55.3          & 50.6          & 29.7         & \multicolumn{1}{c|}{\textbf{22.5}}         & 79.2        & 87.1        & 86.8        & 89.3       & \multicolumn{1}{c|}{\textbf{91.6}}       & 72.3            & 85.6            & 87.0            & 88.0            & \textbf{89.8}            \\ \hline
\multirow{3}{*}{\begin{tabular}[c]{@{}c@{}}SVHN\\ D-BC\end{tabular}}  & LSUN & 22.9          & 22.7          & 22.1          & 18.7         & \multicolumn{1}{c|}{\textbf{16.4}}         & 96.7        & 95.4        & 95.3        & 97.2       & \multicolumn{1}{c|}{\textbf{98.5}}       & 88.0            & 88.7            & \textbf{89.3}            & 86.3            & \textbf{89.3}            \\
                                                                      & C-10 & 30.7          & 20.1          & 24.7          & 20.3         & \multicolumn{1}{c|}{\textbf{12.1}}         & 95.4        & 93.2        & 92.5        & 96.0       & \multicolumn{1}{c|}{\textbf{97.3}}       & 88.5            & 84.7            & 81.7            & 84.2            & \textbf{89.9}            \\
                                                                      & TIN  & 21.2          & 18.6          & 19.9          & 15.2         & \multicolumn{1}{c|}{\textbf{11.7}}         & 97.0        & 96.1        & 95.5        & 97.3       & \multicolumn{1}{c|}{\textbf{98.5}}       & 88.9            & 90.7            & 90.1            & \textbf{91.6}            & 90.6            \\ \hline
\multirow{3}{*}{\begin{tabular}[c]{@{}c@{}}SVHN\\ R-34\end{tabular}}  & LSUN & 25.7          & 21.0          & 22.2          & 18.1         & \multicolumn{1}{c|}{\textbf{13.5}}         & 93.8        & 91.3        & 91.3        & 96.4       & \multicolumn{1}{c|}{\textbf{97.8}}       & 84.6            & 86.5            & 85.9            & 89.4            & \textbf{92.1}            \\
                                                                      & C-10 & 21.7          & 19.5          & 20.0          & 16.7         & \multicolumn{1}{c|}{\textbf{14.8}}         & 94.8        & 92.0        & 91.9        & 97.0       & \multicolumn{1}{c|}{\textbf{97.6}}       & 86.4            & 87.3            & 87.1            & 88.2            & \textbf{89.0}            \\
                                                                      & TIN  & 21.0          & 19.3          & 18.0          & 15.4         & \multicolumn{1}{c|}{\textbf{14.3}}         & 95.4        & 93.4        & 93.5        & 96.8       & \multicolumn{1}{c|}{\textbf{98.2}}       & 86.9            & 88.5            & 88.6            & \textbf{89.4}            & \textbf{89.4}            \\ \bottomrule
\end{tabular}
\end{center}
\end{spacing}
\vskip -0.4in
\end{table}

\vspace{-1.2em}
\subsubsection{Results on CIFAR-10, CIFAR-100 and SVHN.}
We consider sufficient experimental settings in this part for testing the generalization ability of BAL. The pre-trained ResNet-34 and DenseNet-BC on CIFAR-10, CIFAR-100 and SVHN come from \cite{54}. 
The main results on image classification tasks are summarized in Table~\ref{tab:table2}, where BAL demonstrates superior performance compared with the mainstream methods under different experimental settings. Optimal temperature ($T$) and magnitude ($\epsilon$) are searched for ODIN in each group. 
Specifically, BAL reports a decline of FPR95 up to 13.9\% compared with Generalized ODIN.

\vspace{-0.8em}
\subsection{Visualization of \emph{trivial} and \emph{hard OOD} features}
\label{sec:vis}
\vspace{-0.3em}
We show the visualization results of \emph{trivial OOD} features from uniform sampling and the \emph{hard OOD} features from generator via FGSM. We set the training data as two gaussian distribution with dimensionality $m=2$. We use a MLP with three layers as the classifier. The discriminator and generator only use fully connected layers. In the adversarial training process, we sample data in raw data space uniformly since the dimensionality of raw data is fairly low. The other training details are the same as pipeline shown in \textbf{Algorithm~\ref{bal}}.
 
\begin{figure}[htpb]
\centering
\vskip -0.25in
\subfigure[\emph{trivial OOD} via uniform sampling]{
\begin{minipage}[t]{0.48\linewidth}
\centering
\includegraphics[width=2in]{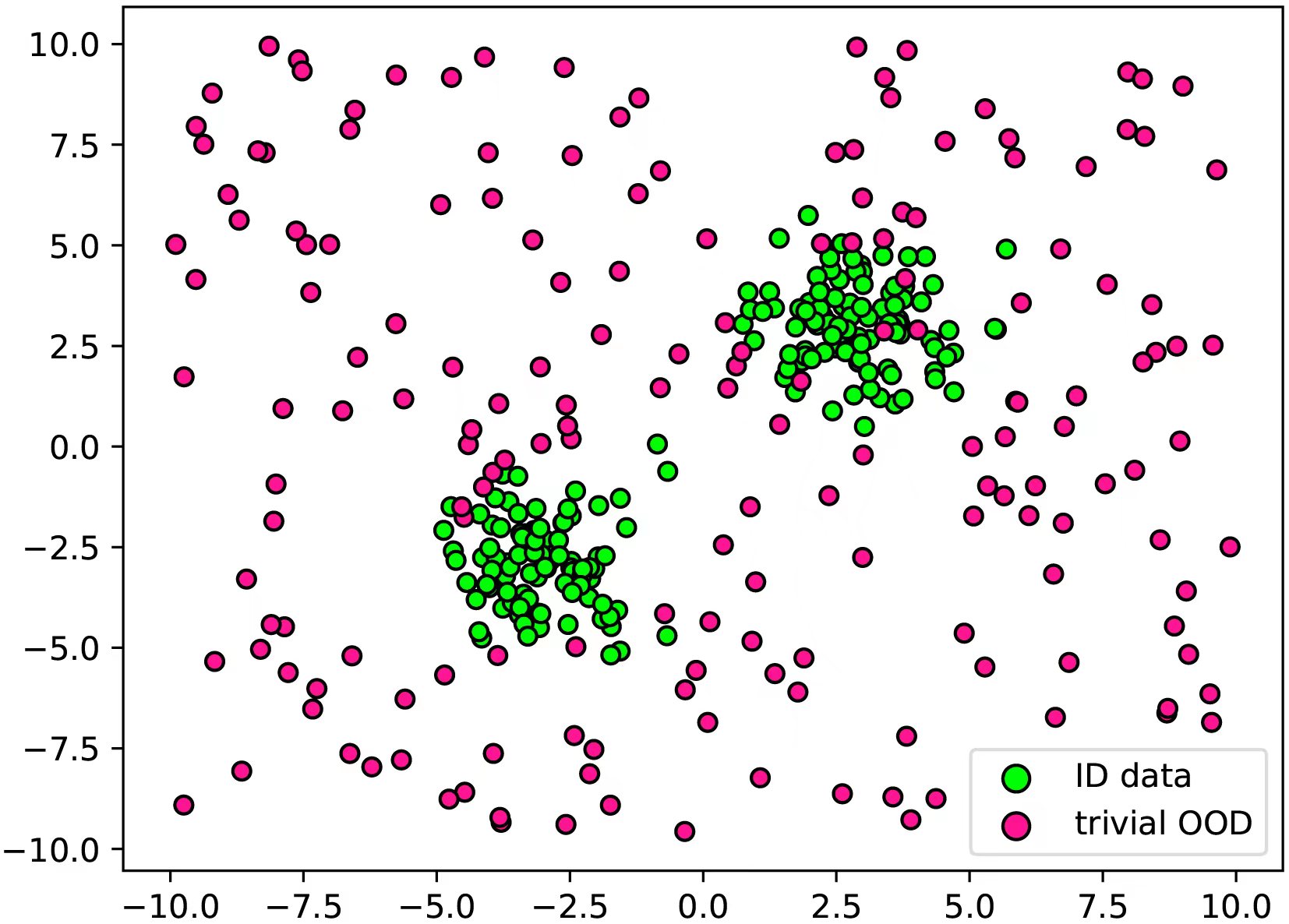}
\end{minipage}%
}%
\subfigure[\emph{hard OOD} via FGSM]{
\begin{minipage}[t]{0.5\linewidth}
\centering
\includegraphics[width=1.98in]{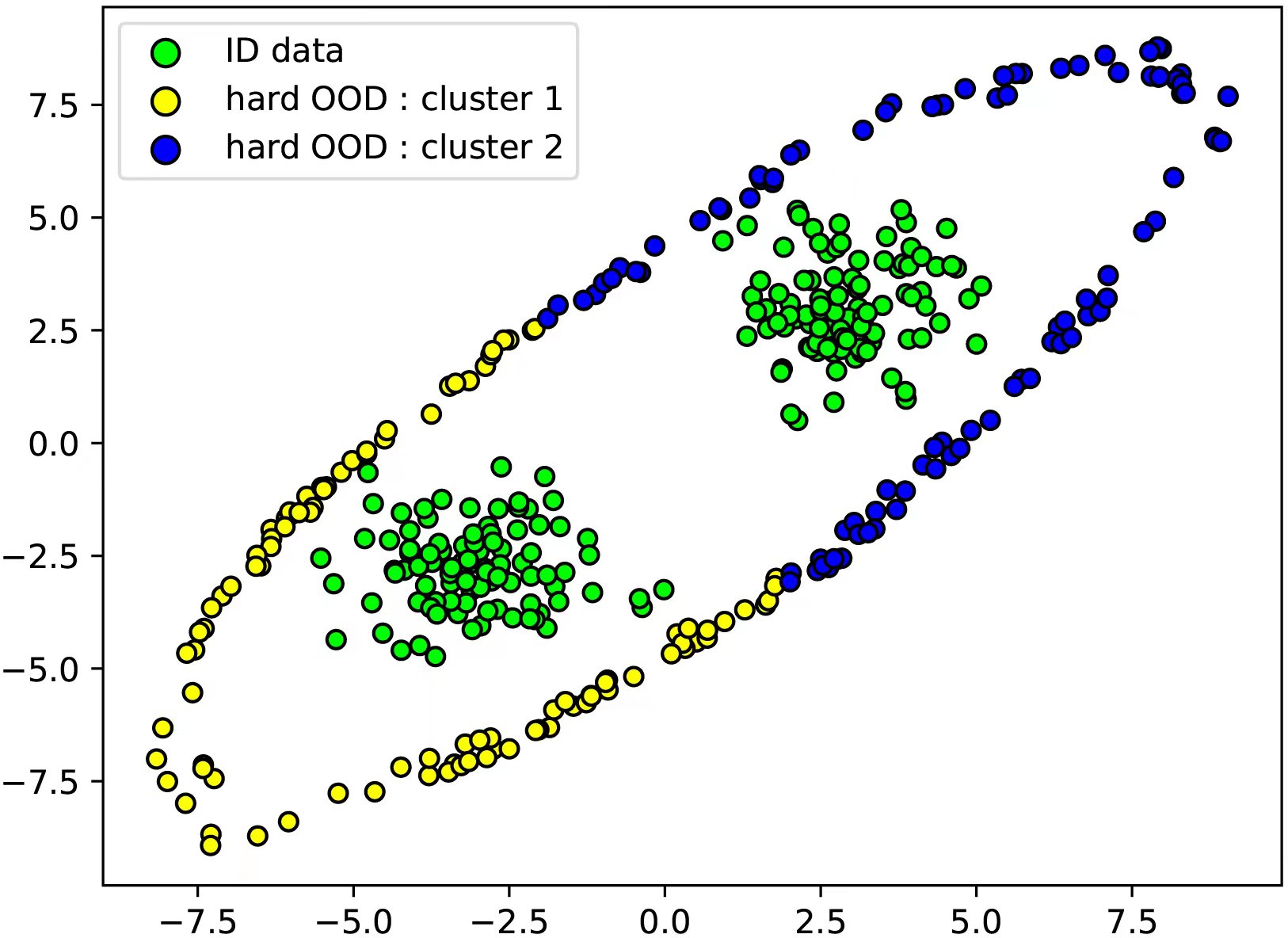}
\end{minipage}%
}%
\centering
\vskip -0.1in
\caption{\textbf{Synthetic \emph{OOD} in raw data space.} When the dimensionality of raw data space is high, we have to perform sampling in feature space as shown in Algorithm~\ref{bal}.}
\label{fig:ood_vis}
\vskip -0.25in
\end{figure}

\begin{figure}[htpb]
\begin{center}
\vskip -0.21in
\includegraphics[width=4.7in]{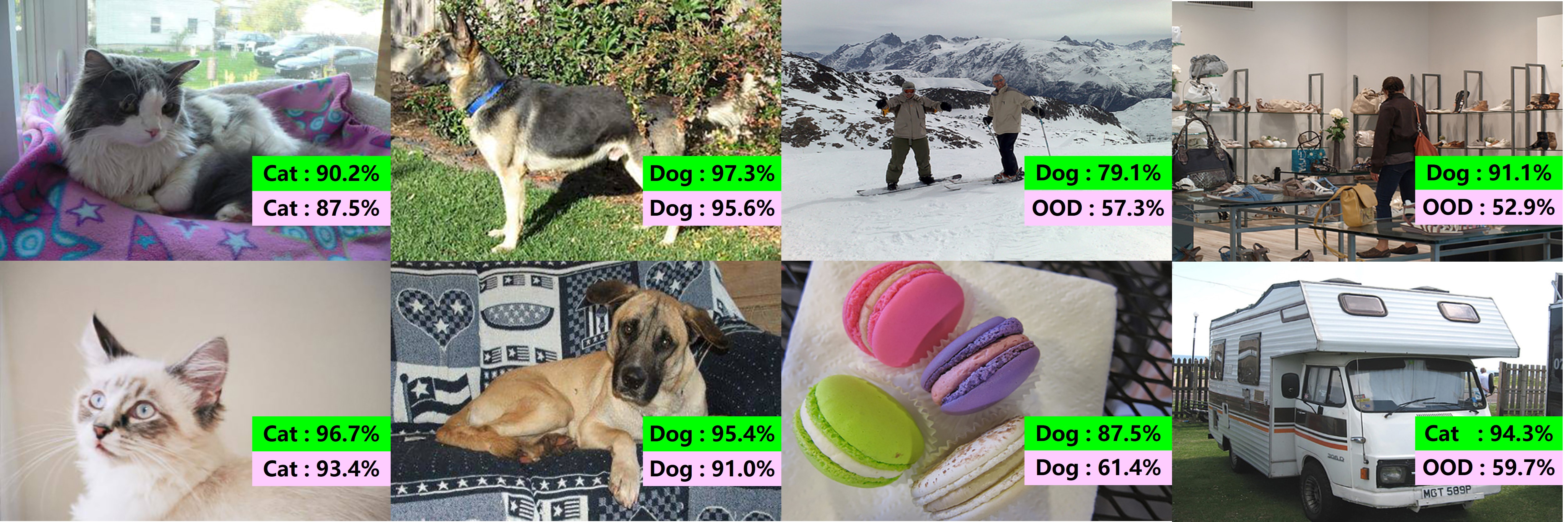}
\vskip -0.1in
\caption{\textbf{\emph{OOD} detection in open world scenario.} Two columns on the left: classification results on \emph{ID} data. Two columns on the right: classification results on \emph{OOD} images from ImageNet. \textbf{Green:} max-softmax baseline. \textbf{Pink:} the proposed BAL. The threshold for distinguish \emph{ID} and \emph{OOD} is set to 0.60 . It is shown that BAL reduces the false positives among classification results. The image with macarons is a failure case where BAL misclassifies it as a dog.}
\label{dog_cat}
\end{center}
\end{figure}

We also report the classification results on dogs vs. cats \cite{53}. The top-1 classification results of BAL and Softmax baseline are given in Figure~\ref{dog_cat}.

\vspace{-4.2em}
\section{Conclusions}
\vspace{-0.7em}
In this paper, we propose \textbf{BAL} to learn the distribution of \emph{OOD} features adaptively. 
No strong assumptions about the \emph{ID} features are introduced. We use a simple uniform sampling method combined with a GAN framework can generate \emph{OOD} features in very high quality step by step. BAL has been proved to generalized well across different datasets and architectures. Experimental results on common image classification benchmarks promise the \emph{state-of-the-art} performance of BAL. The ablation study also shows BAL is stable with different parameter settings. We also report the visualization results of synthetic \emph{OOD} features and open world scenario detection.

\vspace{-1em}
\section*{Acknowledgement}
\vspace{-0.7em}
This research was supported in part by the National Key Research and Development Program of China under Grant No. 2020AAA0109702, and the National Natural Science Foundation of China under Grants 61976208, and the InnoHK project.

\clearpage
\bibliographystyle{splncs04}
\bibliography{ref}
\end{document}